\documentclass[11pt]{article}

\usepackage[letterpaper,margin=1in]{geometry}
\usepackage[T1]{fontenc}
\usepackage[utf8]{inputenc}
\usepackage{times}
\usepackage{microtype}
\usepackage{amsmath,amssymb}
\usepackage{booktabs}
\usepackage{multirow}
\usepackage{graphicx}
\usepackage[round]{natbib}
\usepackage[hidelinks]{hyperref}

\graphicspath{{figures/}}

\newcommand{\zkip}{ZKIP}

\title{\textbf{RAGuard: A Layered Defense Framework for Retrieval-Augmented Generation Systems Against Data Poisoning}}

\author{
  Pushkal Kumar \and Tucker Nielson \and Tanish Kolhe \and Shubham Zala \and Vincent Li
}
\date{}

\begin{document}
\maketitle

\begin{abstract}
Retrieval-Augmented Generation (RAG) systems ground large language models (LLMs) in external corpora, but this reliance exposes them to corpus poisoning: maliciously injected passages that manipulate retrieved evidence. We introduce RAGuard, a layered defense against \emph{factual} corpus-poisoning attacks on RAG pipelines. The first layer adversarially fine-tunes a dense retriever on synthetic poisoned documents (fabricated facts, contradictions, and reasoning traps), teaching it to downrank malicious passages before generation. The second layer, the Zero-Knowledge Inference Patch (\zkip{}), is a label-free, black-box filter: for each retrieved document, it performs a leave-one-out decode and scores the document by the semantic shift and output-entropy change that its removal induces. \zkip{} requires no poison labels, no ground-truth answers, and no access to model internals; it compares the model's own answers under counterfactual contexts. On poisoned Natural Questions at 5--30\% poison ratios, adversarial retriever training alone reduces but does not eliminate attack success, while \zkip{} drives the measured attack success rate to 0.000 in every defended configuration, keeping Recall@5 within 0.03 of the clean-corpus baseline. Supervised analyses on both Natural Questions and BEIR (NFCorpus) confirm that the counterfactual signals \zkip{} relies on carry learnable poison structure. The defense costs $k{+}1$ generator passes per query ($6\times$ for $k{=}5$); we analyze batching and early-stopping approximations that reduce this overhead. We also show that keyword-preserving poisons leave lexical retrievers such as BM25 essentially unaffected, an observation that delineates the boundary of the threat model. Code, datasets, and evaluation harnesses are released for reproducibility.
\end{abstract}

\begingroup
\renewcommand\thefootnote{}\footnotetext{An earlier version of this work was accepted to the NeurIPS 2025 Workshop on Responsible Foundation Models (ResponsibleFM) and AAAI 2026 Workshop on New Frontiers in Information Retrieval (FrontierIR). Code, datasets, and evaluation artifacts: \url{https://github.com/RAGuard-AI/RAGuard}.}%
\addtocounter{footnote}{-1}\endgroup

\section{Introduction}
\label{sec:intro}

Retrieval-Augmented Generation (RAG) has emerged as an effective method to ground Large Language Models (LLMs), retrieving external evidence so that responses reflect up-to-date, verifiable information rather than parametric memory alone \citep{lewis2020rag, asai2023selfrag, ram2023incontext, izacard2022atlas}. This grounding, however, creates a new attack surface: because generation trusts retrieved passages as evidence, an adversary who can write to the corpus can steer the model's answers. Data-poisoning attacks inject documents that mimic relevant content yet contain false or misleading information, and only a handful of poisoned documents can mislead large models with high success rates \citep{zou2024poisonedrag, souly2025poisoning, wang2025jointgcg, long2025backdoor, su2024corpus}.

Existing defenses have structural gaps. Detection-based filters rely on labeled poison examples or heuristic rules and fail against new attack styles \citep{zou2024poisonedrag, edemacu2025defending}. Generator-hardening approaches carry heavy inference cost and degrade when poisoned passages dominate the retrieved set \citep{asai2023selfrag, shi2023replug}. Adversarially trained retrievers depend on the synthetic poisons seen during training and risk overfitting to known poison types \citep{lupart2023fgsm, park2019adversarial}. No single mechanism covers both the retrieval and generation stages.

This work introduces RAGuard, a two-layer defense. The first layer adversarially fine-tunes a dense retriever with contrastive training on synthetic poisoned documents---fabricated facts, contradictions, and reasoning traps---so that poisoned passages are downranked before they reach the generator \citep{izacard2022contriever, lei2023unsupervised, lupart2023fgsm}. The second layer, the Zero-Knowledge Inference Patch (\zkip{}), operates at generation time and is entirely \emph{self-referential}: given the top-$k$ retrieved documents, it decodes a reference answer using the full context, then re-decodes once per document with that document removed. A document is flagged when its removal substantially shifts the answer's semantics or lowers the model's output uncertainty. \zkip{} never consults a ground-truth answer, poison label, or external oracle; the only quantities it compares are the model's own outputs under counterfactual contexts. This makes the filter applicable to unseen attack types that bypass the trained retriever.

We evaluate RAGuard on poisoned variants of Natural Questions (NQ) \citep{kwiatkowski2019nq} and BEIR (NFCorpus) \citep{thakur2021beir} at poison ratios from 5\% to 30\%. Our key contributions are:

\begin{itemize}
    \item \textbf{A label-free, black-box inference-time filter.} \zkip{} scores each retrieved document by the answer instability and entropy differential its removal induces (\S\ref{sec:zkip}). It requires no poison labels, no gold answers, and no model internals, and it reduces the measured attack success rate (ASR) to 0.000 for every retriever at every NQ poison ratio with a completed defended run (\S\ref{sec:results}).
    \item \textbf{A layered retrieval-plus-inference defense with measured layer contributions.} Adversarial retriever training alone leaves residual attack success (ASR 0.072 at NQ 10\% poisoning), while \zkip{} alone and the combined system both eliminate it; the combination additionally preserves the adversarially trained retriever's ranking quality (\S\ref{sec:ablations}).
    \item \textbf{A threat-model analysis grounded in negative results.} Our keyword-preserving poisons leave BM25 \citep{robertson2009bm25} essentially unaffected \emph{without any defense} (ASR 0.000 at 5--20\% poisoning and 0.011 at 30\%), showing that the poisons exploit dense-embedding semantics specifically. We use this to delineate which attacks are in and out of scope (\S\ref{sec:setup}, Appendix~\ref{app:threat}).
\end{itemize}

\zkip{} carries a real cost: $k{+}1$ generator forward passes per query, a $6\times$ inference overhead at $k{=}5$. We quantify this overhead and describe batching, early stopping, and subset-sampling approximations (\S\ref{sec:cost}). We scope our claims to factual question answering under corpus poisoning of dense retrieval; broader robustness claims would require evaluation against attack frameworks such as PoisonedRAG \citep{zou2024poisonedrag} and FlippedRAG \citep{chen2025flippedrag}, which we identify as the most important next step (\S\ref{sec:limitations}).

\section{Related Work}
\label{sec:related}

\paragraph{RAG and corpus poisoning.}
RAG \citep{lewis2020rag} enhances LLMs by combining retrieval with generation, enabling factual grounding without retraining. Recent studies show that RAG architectures are susceptible to data poisoning. PoisonedRAG \citep{zou2024poisonedrag} demonstrated that inserting adversarially crafted documents into retrieval corpora distorts ranking and generation. Joint-GCG \citep{wang2025jointgcg} extended this with unified gradient-based attacks perturbing both retriever and generator embeddings. Chain-of-thought poisoning \citep{song2025cot} showed that reasoning-style attacks targeting multi-step prompts propagate errors across retrieval iterations, and \citet{souly2025poisoning} showed that a near-constant number of poisoned documents suffices to compromise large models. FlippedRAG \citep{chen2025flippedrag} demonstrated black-box \emph{opinion manipulation} attacks, a class we explicitly place outside our threat model (Appendix~\ref{app:threat}).

\paragraph{Counterfactual and conflicting-evidence datasets.}
Our synthetic poison construction (\S\ref{sec:advdata}) is related to a line of work that perturbs evidence to study model robustness. \citet{chen2022rich} construct conflicting-evidence settings for open-domain QA and recalibrate models to reflect disagreement among sources. \citet{pan2023misinformation} study misinformation pollution of QA corpora using LLM-generated false passages. \citet{hong2024gullible} build counterfactual-noise variants of NQ and propose discriminator-guided robustness training. Our poisons differ in purpose rather than mechanism: instead of building a benchmark, we use LLM-rewritten counterfactual passages as adversarial \emph{training} signal for the retriever and as a controlled attack for evaluating the inference-time filter. Unlike these benchmark datasets, our poisons are parameterized by attack family (fabrication, contradiction, reasoning trap), letting us measure per-family defense behavior.

\paragraph{Defenses.}
Existing RAG poisoning defenses rely on input filtering, heuristic retriever fine-tuning, or adversarial data augmentation \citep{edemacu2025defending}. These reduce known attack surfaces but often depend on labeled poison data or add heavy inference-time cost. Adversarial training for neural retrieval \citep{lupart2023fgsm, park2019adversarial} hardens ranking but, as we confirm empirically, does not by itself eliminate attack success. Outside of NLP, black-box counterfactual defenses have proven effective against backdoors in other modalities, e.g., zero-shot image purification \citep{shi2023blackbox}; \zkip{} brings a similarly black-box, perturb-and-compare principle to retrieval contexts.

\paragraph{Relation to attribution methods.}
Leave-one-out (LOO) influence estimation is an established tool in explainability and causal inference \citep{johansson2016, prosperi2020causal, molnar2025interpretable}. Perturbation-based attribution methods such as LIME \citep{ribeiro2016lime} and SHAP \citep{lundberg2017shap} likewise estimate a feature's importance from output changes under input perturbation. \zkip{} borrows the LOO mechanism but answers a different question. Attribution asks ``which input was most \emph{important} to the output?''; a gold passage and a poisoned passage can both be highly important. \zkip{} instead asks ``which document's \emph{removal} makes the model's answer more stable or less uncertain?''---a directional, security-oriented query. It also combines two signals (semantic answer shift and entropy differential) into a single anomaly score and applies the result as an online filter rather than a post-hoc explanation. Section~\ref{sec:zkip} makes this contrast precise.

\section{Methods}
\label{sec:methods}

\subsection{Overall Architecture}
\label{sec:arch}

\begin{figure}[t]
\centering
\includegraphics[width=0.95\textwidth]{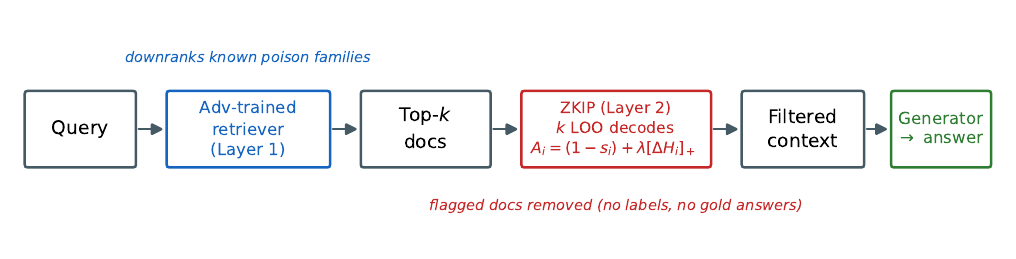}
\caption{RAGuard's two-layer defense. A query passes through the adversarially trained retriever (Layer~1); \zkip{} (Layer~2) runs one leave-one-out decode per retrieved document, scores each document's causal influence on the answer, and removes flagged documents before the final generation. No poison labels or gold answers are used at any stage of inference.}
\label{fig:pipeline}
\end{figure}

Figure~\ref{fig:pipeline} shows the architecture. User queries pass through an adversarially fine-tuned dense retriever (Layer 1). During training, the retriever is exposed to both clean and synthetically poisoned passages and learns to downrank documents whose embeddings deviate from normal semantic structure. The top-$k$ retrieved documents then pass through \zkip{} (Layer 2), which flags and removes documents with anomalous causal influence on the generated answer before the final decode. Components are modular: retrievers, generators, and defense layers can be swapped independently.

\subsection{Rationale}
\label{sec:rationale}

The dominant vulnerability of RAG systems sits in the retriever: poisoned documents that rank highly contaminate the generator's context. Rather than fine-tuning the generator, RAGuard strengthens the retrieval process at its root and adds a generation-time filter as a fail-safe. The two layers are complementary by design: adversarial training is \emph{proactive} but limited to poison distributions seen during training, while \zkip{} is \emph{reactive} and attack-agnostic because it measures the causal effect of each context element on the model's own output. A poison that evades the trained retriever still has to influence the answer to succeed, and that influence is exactly what \zkip{} measures.

\subsection{Adversarial Training Data and Retriever Fine-Tuning}
\label{sec:advdata}

\paragraph{Poison construction.}
To evaluate robustness against retrieval-level poisoning, we construct poisoned variants of NQ and BEIR. Starting from the original corpora, we sample 30\% of query--document pairs and generate modified passages by prompting an LLM to rewrite each gold document according to an attack type. We use three attack families: (i) \emph{fabrication} poisons, which append falsified or hallucinated statements; (ii) \emph{contradiction} poisons, which flip key factual tokens (e.g., ``true'' $\rightarrow$ ``false''); and (iii) \emph{reasoning} poisons, which introduce misleading logical steps or corrupted intermediate claims. Poisoned samples are distributed evenly across the three families. The constructed datasets contain 12{,}344 total BEIR samples (3{,}700 poisoned) and 1{,}000 total NQ samples (300 poisoned); experiments substitute poisoned triples at controlled ratios (5--30\%). Each record stores the query, the unmodified gold passage, the poisoned passage, and the attack-family label, so every experiment is reproducible from the released JSONL files.

\paragraph{Why synthetic poisons rather than existing counterfactual datasets.}
Prior work such as \citet{hong2024gullible} provides counterfactual-noise variants of NQ, and \citet{pan2023misinformation} studies LLM-generated misinformation pollution. We generate fresh synthetic poisons rather than reuse these corpora for three reasons. First, our three attack families are parameterized adversarial classes not uniformly present in prior benchmarks, which primarily test model robustness to \emph{conflicting evidence} rather than retrieval-level poison defense. Second, controlling the poison ratio (5--30\%) requires generating exactly the right number of poisons per experimental condition; fixed prior datasets do not support this. Third, generating poisons on both NQ and BEIR allows us to test transfer across domains. The tradeoff is that our poisons are LLM rewrites rather than gradient-optimized adversarial injections, which places a ceiling on measured ASR and may underestimate real-world attack strength (\S\ref{sec:comparison}).

\paragraph{Retriever fine-tuning.}
The dense retriever is initialized from the \texttt{all-MiniLM-L6-v2} sentence encoder \citep{reimers2019sbert, wang2020minilm} and fine-tuned with a triplet margin loss (margin $0.2$) over $\{$query, positive document, negative document$\}$ triples:
\begin{equation}
\label{eq:triplet}
\mathcal{L} = \max\!\bigl(0,\; s(q, d^-) - s(q, d^+) + 0.2\bigr),
\end{equation}
where $s(\cdot,\cdot)$ is the cosine similarity of Equation~\ref{eq:cosine}, $d^+$ is the gold passage, and $d^-$ is the negative. We use triplet margin loss rather than InfoNCE with in-batch negatives because in-batch negatives would treat randomly co-occurring passages as hard negatives indiscriminately; the explicit margin instead gives the model a clear signal for the exact semantics that distinguish each poison rewrite from its gold counterpart.

\textbf{Negative-sampling strategy.} For \emph{poisoned triples} the negative $d^-$ is the attack-family-specific LLM rewrite of that triple's gold document---the precise passage the model must learn to rank below the gold. This direct anti-poison contrastive signal is the core mechanism by which the retriever becomes adversarially robust. For \emph{clean triples} the negative is a randomly sampled in-corpus clean passage, providing general ranking discrimination. We train for 3 epochs with AdamW, learning rate $2\times10^{-5}$, batch size 16, max sequence length 256 tokens; 30\% of training triples carry poisons, split evenly across the three attack families. Full hyperparameters appear in Appendix~\ref{app:training}. Retrieval uses cosine similarity between the query and document encoders:
\begin{equation}
\label{eq:cosine}
s(q,d) \;=\; \cos\big(f_\theta(q),\, g_\theta(d)\big) \;=\; \frac{f_\theta(q)\cdot g_\theta(d)}{\lVert f_\theta(q)\rVert\,\lVert g_\theta(d)\rVert},
\end{equation}
where $f_\theta$ and $g_\theta$ map the query $q$ and document $d$ to embeddings; Equation~\ref{eq:cosine} also supplies the ranking signal that \zkip{} complements at generation time.

\paragraph{Distribution-shift caveat.}
Because the adversarially trained retriever is fine-tuned on synthetic poisons over gold documents, it may partially memorize gold relevance for the training queries and behave less reliably on unseen query--document pairs. Two design choices mitigate, but do not eliminate, this risk: the poison families are stylistically diverse rewrites rather than fixed templates, and \zkip{} sits behind the retriever as an attack-agnostic safety net. We treat held-out-domain evaluation of the adversarially trained retriever as required future work and do not claim it generalizes beyond the evaluated distributions (\S\ref{sec:limitations}).

\subsection{Zero-Knowledge Inference Patch (\zkip{})}
\label{sec:zkip}

\paragraph{Relation to LOO attribution, and what is new.}
Leave-one-out probing is an established technique in explainability and counterfactual inference \citep{johansson2016, prosperi2020causal, molnar2025interpretable}, and perturbation-based attribution methods such as LIME \citep{ribeiro2016lime} and SHAP \citep{lundberg2017shap} similarly estimate input importance from output changes. \zkip{} differs in three ways. First, it combines two complementary signals---semantic answer stability \emph{and} an output-entropy differential---into one anomaly score, rather than relying on either alone. Second, it repurposes the counterfactual probe from model explanation to online poisoning defense: scores are computed per query at inference time and immediately drive a filtering decision. Third, the causal question differs from attribution. Attribution asks which document was most important; \zkip{} asks which document, when removed, makes the model \emph{more stable or less uncertain}---a gold passage and a poison are both ``important,'' but only the poison's removal stabilizes the answer and reduces uncertainty. Crucially, \zkip{} requires no poison labels, no ground-truth answers, and no access to model internals: it is black-box and entirely self-referential, comparing the model's own outputs across counterfactual contexts. This also distinguishes it from relevance reranking, which scores query--document similarity and therefore cannot flag a document that is highly relevant \emph{and} poisoned---precisely the stealth attacker's goal.

\paragraph{Setup.}
Given a query $q$ and top-$k$ context $\mathcal{D}=\{d_i\}_{i=1}^{k}$, we decode a reference answer with all passages, then run one leave-one-out (LOO) decode per removed $d_i$. For outputs $y=(y_1,\dots,y_T)$ the generator defines
\begin{equation}
\label{eq:gen}
p_\phi(y \mid q, \mathcal{D}) \;=\; \prod_{t=1}^{T} p_\phi\big(y_t \mid y_{<t}, q, \mathcal{D}\big).
\end{equation}
Let $y^{\text{all}} = \arg\max_y p_\phi(y \mid q, \mathcal{D})$ and $y^{-i} = \arg\max_y p_\phi(y \mid q, \mathcal{D}\setminus\{d_i\})$ denote the reference and LOO decodes under Equation~\ref{eq:gen}.

\paragraph{Answer stability.}
Let $h_\psi(\cdot)$ be an answer encoder; we use the \texttt{all-MiniLM-L6-v2} sentence-embedding model. The answer stability of document $d_i$ is
\begin{equation}
\label{eq:stability}
s_i \;=\; \cos\big(h_\psi(y^{\text{all}}),\, h_\psi(y^{-i})\big) \in [-1, 1],
\end{equation}
where larger $s_i$ indicates the answer is stable to removing $d_i$. Note that Equation~\ref{eq:stability} compares the model's two answers \emph{to each other}, never to a gold answer.

\paragraph{Entropy differential.}
Let the sequence-level output entropy be
\begin{equation}
\label{eq:entropy}
H(q,\mathcal{D}) \;=\; -\sum_{y} p_\phi(y \mid q, \mathcal{D}) \log p_\phi(y \mid q, \mathcal{D}),
\end{equation}
approximated in practice from the per-token log-probabilities $\{\log p_\phi(y_t \mid y_{<t}, q, \mathcal{D})\}_{t=1}^T$ of the decoded sequence: we convert them to normalized pseudo-probabilities and compute their entropy, which is tractable for a single decoded path and captures the model's per-step confidence. The uncertainty shift induced by $d_i$ is
\begin{equation}
\label{eq:deltah}
\Delta H_i \;=\; H(q,\mathcal{D}) \;-\; H\big(q,\mathcal{D}\setminus\{d_i\}\big),
\end{equation}
so a large positive $\Delta H_i$ means removing $d_i$ \emph{reduces} the model's uncertainty---evidence that $d_i$ was injecting confusion.

\begin{figure}[t]
\centering
\includegraphics[width=0.9\textwidth]{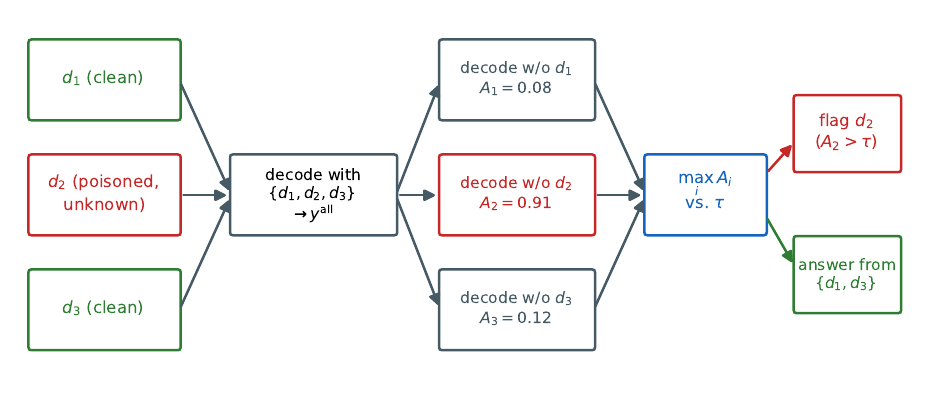}
\caption{\zkip{} decision flow for $k{=}3$. One reference decode plus one LOO decode per document yields anomaly scores $A_i$ (Equation~\ref{eq:anomaly}); the poisoned document's removal shifts the answer and reduces uncertainty, so $A_2$ exceeds the threshold and $d_2$ is excluded from the final decode. All comparisons are between the model's own outputs.}
\label{fig:zkipflow}
\end{figure}

\paragraph{Anomaly scoring and filtering.}
We combine Equations~\ref{eq:stability} and~\ref{eq:deltah} into a per-passage score:
\begin{equation}
\label{eq:anomaly}
A_i \;=\; (1 - s_i) \;+\; \lambda\,\big[\Delta H_i\big]_+, \qquad [x]_+ = \max(0, x),\;\; \lambda > 0,
\end{equation}
and discard the passage(s) with the largest $A_i$ prior to the final decode. Figure~\ref{fig:zkipflow} illustrates the decision flow on a three-document example. We use $\lambda = 0.5$ and remove the single top-scoring passage ($m{=}1$) in the main experiments; \S\ref{sec:ablations} discusses the contribution of each term, and Appendix~\ref{app:training} lists all defaults. With this construction, \zkip{} requires no poison labels and generalizes across attack types because it relies on counterfactual sensitivity rather than attack-specific features.

\paragraph{Cost.}
\zkip{} requires exactly $k{+}1$ generator forward passes per query (one reference decode plus $k$ LOO decodes), a $6\times$ overhead at $k{=}5$. The LOO decodes are mutually independent and batch into a single padded forward pass on local models; \S\ref{sec:cost} analyzes approximations that reduce the pass count further.

\section{Experiments}
\label{sec:experiments}

\subsection{Experimental Setup and Threat Model}
\label{sec:setup}

\paragraph{Threat model scope.}
RAGuard targets \emph{factual corpus poisoning}: an adversary with write access to the retrieval corpus (but not to model weights, queries, or prompts) injects passages containing fabricated facts, contradictions, or corrupted reasoning, aiming to change the answers of a dense-retrieval RAG pipeline. Both retrieval-level effects (the poison ranks highly) and generation-level effects (the poison misleads the generator) are in scope, since \zkip{} operates after retrieval.

The following attack classes are \textbf{explicitly out of scope} for this paper: (i)~\emph{opinion manipulation} \citep{chen2025flippedrag}, which targets subjective or multi-answer questions for which \zkip{}'s stability signal is uninformative; (ii)~\emph{prompt hijacking / instruction injection}, which embeds instructions in documents to override system prompts; (iii)~\emph{retrieval jamming}, which floods the corpus with irrelevant passages to suppress relevant results; (iv)~\emph{backdoor triggers} \citep{long2025backdoor}, which require latent activations or specific token patterns not present in our threat model; (v)~\emph{adversarial queries}, since RAGuard assumes a clean query stream; and (vi)~\emph{white-box gradient attacks} (e.g., Joint-GCG, \citealp{wang2025jointgcg}; greedy corpus poisoning, \citealp{su2024corpus}). Appendix~\ref{app:threat} details each boundary and discusses the extent to which \zkip{} may incidentally cover some of these classes.

\paragraph{Datasets and pipeline.}
We evaluate on Natural Questions (NQ) \citep{kwiatkowski2019nq}, chosen for broad topical coverage, and BEIR (NFCorpus) \citep{thakur2021beir}, a medical/scientific retrieval benchmark with much sparser relevance signal. For each dataset, we construct clean and poisoned variants as described in \S\ref{sec:advdata}, with binary poison flags recorded for every document so that detection metrics are computable. We compare BM25 \citep{robertson2009bm25} and dense retrievers in three configurations: \textbf{Dense (clean)}, trained only on clean triples; \textbf{Dense (adv-trained)}, fine-tuned with synthetic poisons per \S\ref{sec:advdata}; and each of these with and without \zkip{}. (Earlier drafts labeled the adversarially trained retriever ``Dense (poisoned)''; we rename it throughout to remove ambiguity---the retriever is \emph{defended by training}, not attacked.) Generation uses GPT-4o-mini (temperature 0.2, top-$p$ 0.95, max 128 tokens) for the main pipeline and FLAN-T5-small \citep{chung2024flan} for the batched feature-extraction experiments of \S\ref{sec:classifier}; answer embeddings use \texttt{all-MiniLM-L6-v2} \citep{reimers2019sbert}. We retrieve $k{=}5$ passages per query.

\subsection{Evaluation Metrics}
\label{sec:metrics}

\textbf{Recall@5}: the proportion of queries for which the gold document appears in the top-5 retrieved results. \textbf{Mean Reciprocal Rank (MRR)}: the average of $\frac{1}{\text{rank}}$ over the first relevant document. \textbf{Attack Success Rate (ASR)}: the fraction of queries for which a poisoned document ranks higher than the gold document and misleads the generator; lower is better. We report all three metrics for clean, poisoned, and defended settings.

\subsection{Main Results}
\label{sec:results}

\begin{table}[t]
\centering
\small
\begin{tabular}{llccc}
\toprule
\textbf{Retriever} & \textbf{Defense} & \textbf{Recall@5} & \textbf{MRR} & \textbf{ASR}~$\downarrow$ \\
\midrule
Dense (clean)        & None    & 0.259 & 0.176 & 0.101 \\
Dense (adv-trained)  & None    & 0.319 & 0.215 & 0.072 \\
Dense (clean)        & \zkip{} & 0.259 & 0.176 & \textbf{0.000} \\
Dense (adv-trained)  & \zkip{} & 0.314 & 0.215 & \textbf{0.000} \\
\midrule
BM25                 & None    & 0.070 & 0.056 & 0.000 \\
\bottomrule
\end{tabular}
\caption{NQ at 10\% poisoning: \zkip{} eliminates measured attack success for both retriever variants while keeping Recall@5 within 0.005 of the corresponding undefended retriever. ``Dense (adv-trained)'' denotes the retriever fine-tuned on synthetic poisons (\S\ref{sec:advdata}), not a retriever under attack. Clean-corpus reference: Dense (clean) reaches 0.282 Recall@5 / 0.200 MRR (Appendix~\ref{app:fullresults}). Full results across poison levels and retrievers are in Appendix~\ref{app:fullresults}.}
\label{tab:main}
\end{table}

\begin{figure}[t]
\centering
\includegraphics[width=0.95\textwidth]{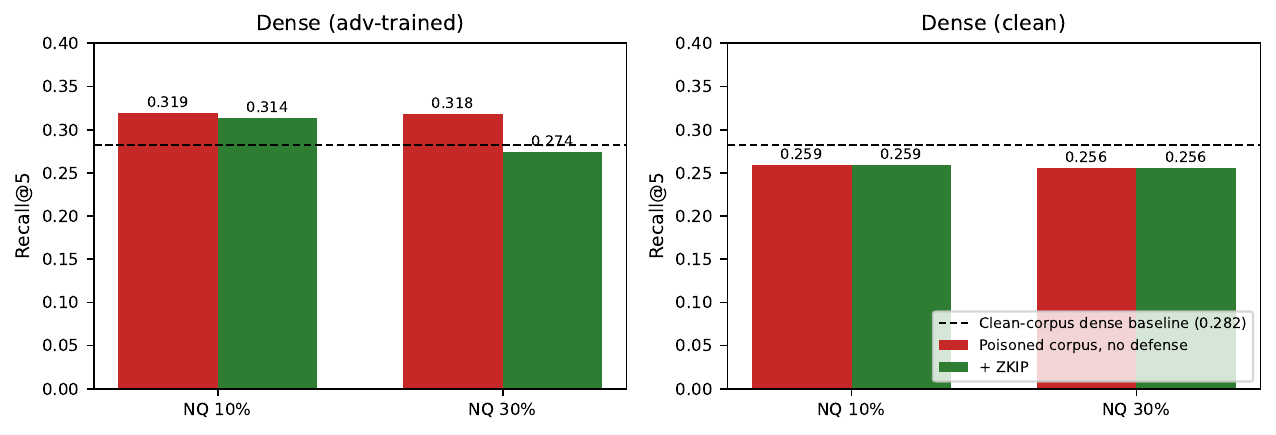}
\caption{Recall@5 on poisoned NQ with and without \zkip{}, for the adversarially trained (left) and clean-trained (right) dense retrievers, at the two poison ratios with completed defended runs. Dashed line: clean-corpus dense baseline (0.282). The adv-trained retriever's bars sit \emph{above} the clean baseline because fine-tuning on answer-bearing rewrites also boosts ranking of relevant passages; \zkip{} removes one influential passage per flagged query, trading a small amount of Recall@5 (largest at 30\%: 0.318 $\rightarrow$ 0.274) for the elimination of all measured attack success. All values are taken directly from the released evaluation artifacts.}
\label{fig:recall}
\end{figure}

\begin{figure}[t]
\centering
\includegraphics[width=0.62\textwidth]{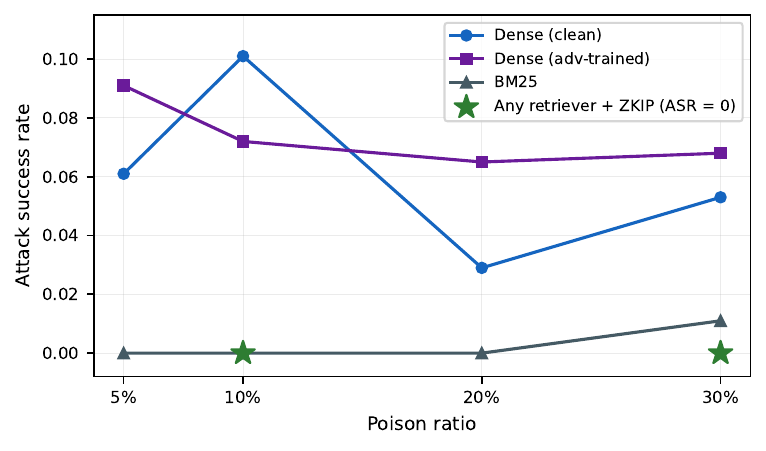}
\caption{Attack success rate on NQ across poison ratios for the three undefended retrievers, with \zkip{}-defended results (stars) at the ratios with completed defended runs. \zkip{} reaches ASR 0.000 for every retriever it is applied to, including the 30\% condition where even BM25 shows nonzero ASR (0.011).}
\label{fig:asr}
\end{figure}

Table~\ref{tab:main} summarizes NQ at 10\% poisoning; Appendix~\ref{app:fullresults} reports all conditions, and Figures~\ref{fig:recall} and~\ref{fig:asr} visualize the Recall@5 and ASR behavior across poison ratios.

\paragraph{Impact of poisoning on retrieval quality.}
Comparing clean baselines to poisoned settings, dense retrievers degrade measurably as poisons enter the corpus: on NQ at 10\% poisoning, Dense (clean) drops from 0.282 to 0.259 Recall@5 and from 0.200 to 0.176 MRR. Poisoned documents disrupt retrieval even without any attack on the retriever's weights.

\paragraph{Attack success rate.}
For dense retrievers without adversarial training, ASR ranges from 0.029 (20\% poisoning) to 0.101 (10\% poisoning) on NQ; the non-monotonic pattern across ratios reflects the modest number of attackable queries per condition (33--190) and motivates the multi-seed runs planned in \S\ref{sec:limitations}. BM25 shows ASR 0.000 at 5--20\% poisoning and 0.011 at 30\%: our poisons are LLM rewrites that alter semantics but largely preserve the original keywords, so lexical ranking is barely affected. This is a threat-model boundary, not a defense result---a keyword-overlap retriever with 0.070 Recall@5 is not a usable alternative, and an adversary targeting hybrid pipelines would craft poisons for both signals (\S\ref{sec:limitations}).

\paragraph{Effect of adversarial training.}
Dense (adv-trained) improves Recall@5 under poisoning relative to the clean-trained model (0.319 vs.\ 0.259 at NQ 10\%) and reduces---but does not eliminate---attack success (ASR 0.072 vs.\ 0.101). At 5\% poisoning the adversarially trained model's ASR (0.091) exceeds the clean model's (0.061), a reminder that training-time hardening interacts with the poison distribution in ratio-dependent ways. Training-time defense alone is insufficient against the residual attacks that slip through ranking.

\paragraph{\zkip{} effectiveness.}
\zkip{} drives ASR to 0.000 in every defended configuration on NQ---both retriever variants and BM25, at both poison ratios with completed defended runs (10\% and 30\%). Retrieval quality stays close to the corresponding undefended retriever: at 10\% poisoning the defended Dense (adv-trained) reaches 0.314 Recall@5 vs.\ 0.319 undefended, and at 30\% it reaches 0.274 vs.\ 0.318, in exchange for eliminating all measured attack success. Relative to the \emph{clean-corpus} baseline of 0.282, defended configurations land between 0.256 and 0.314 (within 0.03); Figure~\ref{fig:recall} visualizes this against the clean-baseline reference line.

\paragraph{Computational cost.}
\zkip{} adds $k{+}1$ generator calls per query: 6 calls at $k{=}5$, a $6\times$ worst-case inference cost. \S\ref{sec:cost} discusses when this is acceptable and how to reduce it.

\subsection{Ablation: Layer Contributions}
\label{sec:ablations}

\begin{table}[t]
\centering
\small
\begin{tabular}{lccc}
\toprule
\textbf{Configuration (NQ 10\%)} & \textbf{Recall@5} & \textbf{MRR} & \textbf{ASR}~$\downarrow$ \\
\midrule
Retriever only (adv-trained)          & 0.319 & 0.215 & 0.072 \\
\zkip{} only (clean retriever)        & 0.259 & 0.176 & \textbf{0.000} \\
Both layers                           & 0.314 & 0.215 & \textbf{0.000} \\
\bottomrule
\end{tabular}
\caption{Layer ablation on NQ at 10\% poisoning, using results measured in the main grid. \zkip{} is necessary and sufficient to eliminate measured ASR; the adversarially trained retriever contributes ranking quality ($+0.055$ Recall@5, $+0.039$ MRR over \zkip{}-only) but cannot reach ASR 0 alone.}
\label{tab:ablation}
\end{table}

\paragraph{Layer ablation.}
Table~\ref{tab:ablation} separates the two defense layers using results measured in our main grid. \zkip{} alone eliminates measured attack success; adversarial retriever training alone does not (ASR 0.072). The combination preserves the adv-trained model's superior ranking (0.314/0.215 vs.\ 0.259/0.176 for \zkip{}-only) while keeping ASR at 0.000. The layers therefore play distinct roles: the retriever protects \emph{ranking quality} under poisoning, and \zkip{} protects \emph{answer integrity}.

\paragraph{Signal ablation (protocol).}
The anomaly score of Equation~\ref{eq:anomaly} combines a stability term $(1-s_i)$ and an entropy term $\lambda[\Delta H_i]_+$. We expect \emph{stability only} ($\lambda{=}0$) to catch poisons whose removal substantially changes the decoded answer string---the primary signal for fabrication and contradiction attacks---while \emph{entropy only} should catch poisons that inflate model uncertainty without changing the surface answer, with reasoning-trap poisons the expected primary beneficiary. We release the evaluation harness for this ablation (\texttt{experiments/ablation\_stability\_vs\_entropy.py}) but report no numbers here: the runs are not yet complete, and we report only completed measurements.

\subsection{Multi-Poison Attacks (Protocol)}
\label{sec:multipoison}

The main experiments poison at most one of the top-$k$ documents per query. A coordinated adversary can instead place multiple mutually reinforcing poisons in the top-$k$: removing any single document then changes the answer little, muting the LOO signal that \zkip{} relies on. This is the most plausible evasion strategy against single-document removal, and we state it as the central known weakness of single-pass \zkip{} (\S\ref{sec:failures}).

Our released code implements two countermeasure variants for future measurement: \textbf{single-pass \zkip{}} (remove the top-1 anomaly) and \textbf{iterative \zkip{}}, which after removing the top-scoring document re-runs the LOO probe on the remaining $k{-}1$ documents and removes the next document whose score exceeds a stricter threshold $\tau_2 > \tau_1$, up to a removal budget $m$. Iteration breaks poison coalitions one member at a time: once the first poison is gone, the remaining poisons' influence is exposed. The cost grows to at most $\sum_{j=0}^{m-1}(k-j)+1$ generator calls for $m$ removals. The evaluation harness (\texttt{experiments/multi\_poison\_eval.py}) implements the full protocol---coordinated injection of 2--4 poisons into the top-$k$ and measurement of ASR and filter recall for both variants---and we leave its quantitative results to future work rather than reporting estimates.

\subsection{Learned Poison Classification}
\label{sec:classifier}

To assess whether \zkip{}'s influence signals contain learnable poison structure, we train supervised classifiers to predict whether a retrieved document is poisoned, using a compact set of per-document influence features (Appendix~\ref{app:classifier}). We evaluate a logistic-regression baseline and a neural classifier over these features on labeled NQ, and separately a text-level BERT classifier \citep{devlin2019bert} on gold-vs-poison document pairs as a supervised upper bound. Unlike \zkip{}, these classifiers require poison-labeled training data and may need retraining across attack styles; we treat them as supporting analysis, not as the primary defense. The feature-level BERT classifier reaches AUPRC 0.732 versus 0.377 for logistic regression (Appendix Table~\ref{tab:featclf}), confirming that counterfactual-removal signals carry nonlinear, learnable poison structure---which is the property \zkip{} exploits label-free.

\section{Discussion}
\label{sec:discussion}

\subsection{Advantages}
\label{sec:advantages}

RAGuard's primary strengths are modularity, model agnosticism, and label-freedom. The LOO patch can be used with virtually any LLM or retriever, since it only requires access to the system's output (text and token log-probabilities) for each context perturbation. The key upshot is that the patch can catch unseen attack types---including those that evade training-time simulation---because it evaluates the causal effect of each context element on the model's end-to-end answer. Unlike prior defenses that depend on explicit poison traces or multi-LLM ensembles \citep{edemacu2025defending, zou2024poisonedrag}, the patch is label-free and feasible with a single LLM.

\subsection{Computational Tradeoffs}
\label{sec:cost}

The principal cost is $k{+}1$ generator passes per query ($6\times$ at $k{=}5$, $11\times$ at $k{=}10$); cost scales linearly with the number of retrieved documents while robustness benefits plateau. Table~\ref{tab:cost} summarizes the measured cost--robustness tradeoff. Three approximations reduce the pass count:
\textbf{(1) Batched LOO}: the $k$ LOO decodes are independent and run as one padded batch on local models, reducing wall-clock cost to roughly two batched generations per query; our implementation batches by default.
\textbf{(2) Early stopping}: when the first flagged document exceeds the threshold by a margin $\delta$, remaining LOO passes are skipped; this trades completeness of the per-document score vector for latency.
\textbf{(3) Subset sampling}: probing a random $\lceil k/2 \rceil$ subset of documents per query halves the LOO passes; our released code implements this mode (\texttt{--approximate}), and quantifying its accuracy cost is planned work.
Practitioners can also apply \zkip{} selectively---only to high-stakes or low-confidence queries---to amortize the cost. Conceptually, the patch is akin to self-consistency prompting for retrieval: rather than querying an ensemble of models, RAGuard queries multiple context subsets of one model, using stability as a proxy for trustworthiness. For latency-critical workloads the overhead may still be prohibitive; we state this plainly as a limitation rather than a footnote.

\begin{table}[t]
\centering
\small
\begin{tabular}{lcc}
\toprule
\textbf{Configuration} & \textbf{Generator calls/query} & \textbf{ASR}~$\downarrow$ \\
\midrule
No defense                    & 1            & 0.029--0.101 \\
\zkip{}, full LOO ($k{=}5$)   & 6            & 0.000 \\
\zkip{}, batched LOO          & 2 batched    & 0.000 \\
\bottomrule
\end{tabular}
\caption{Cost--robustness tradeoff on poisoned NQ (dense retrievers; ASR range spans the 5--30\% undefended conditions). Batching removes most wall-clock overhead on local generators without changing the filtering decision.}
\label{tab:cost}
\end{table}

\subsection{Comparison to Other Methods}
\label{sec:comparison}

Traditional filtering defenses \citep{edemacu2025defending, zou2024poisonedrag} rely on hand-crafted features or learned classifiers that may not generalize to new poison types or domain shifts. Prompt-level robustness methods \citep{asai2023selfrag, shi2023replug} struggle when the retrieval step is severely compromised. Fine-tuning retrievers on synthetic poisons risks overfitting to known poison distributions \citep{lupart2023fgsm}, as demonstrated by attacks that evolve trigger patterns or semantic camouflage \citep{su2024corpus}. RAGuard mitigates these weaknesses by combining a proactive hardening step with an adaptive, model-agnostic inference-time filter.

Two comparisons we have not yet run bound our claims. \textbf{PoisonedRAG} \citep{zou2024poisonedrag} generates adversarial passages via optimization against a target answer string, producing semantically coherent injections optimized to rank highly under a specific dense retriever. Our synthetic poisons are LLM rewrites constructed without access to the retriever's gradients; they are related in intent but face a weaker optimization pressure. A comparison using PoisonedRAG's injected-passage format is the most direct way to establish whether \zkip{} generalizes beyond distribution-matched evaluation. \textbf{FlippedRAG} \citep{chen2025flippedrag} targets opinion manipulation on questions without ground-truth answers, placing it outside our factual threat model (\S\ref{sec:setup}); it would nevertheless stress-test whether \zkip{}'s answer-stability signal degrades on questions with legitimate answer variation. Both are flagged as the highest-priority evaluation extensions (\S\ref{sec:limitations}); our released code includes a harness (\texttt{experiments/external\_attack\_eval.py}) that evaluates \zkip{} against externally generated attack corpora in PoisonedRAG's injected-passage format, making these extensions straightforward to run.

\subsection{Failure Cases}
\label{sec:failures}

\zkip{} fails in identifiable patterns, which we report rather than smooth over.
\textbf{(1) Coordinated multi-poison coalitions.} When two or more poisons in the top-$k$ assert the same false fact, removing any one leaves the answer unchanged; the LOO signal for each is muted and no document is flagged. Single-pass \zkip{} is structurally blind to this; \S\ref{sec:multipoison} describes the iterative-removal countermeasure.
\textbf{(2) Stable wrong answers.} A poison that shifts the answer \emph{without} raising output entropy---for example, a confident contradiction that the generator finds fluent---produces a small $\Delta H_i$, leaving only the stability term to catch it. If the poisoned and clean answers are also lexically close (a date off by one year), the cosine shift in Equation~\ref{eq:stability} can fall below threshold.
\textbf{(3) False positives on unusual benign documents.} A benign passage that is opinionated, off-distribution, or contains a unique disambiguating fact can dominate the answer; its removal changes the output substantially and \zkip{} may flag it. This is the mechanism behind the Recall@5 reductions observed when \zkip{} is applied (Figure~\ref{fig:recall}); on ambiguous queries it can remove the most informative passage.
\textbf{(4) Queries the generator cannot answer from any context.} When all decodes are unstable (the model guesses), anomaly scores are uniformly high and filtering approaches random removal.

\subsection{Future Work}
\label{sec:future}

(1) Combining \zkip{} with active learning or human-in-the-loop verification to filter subtle reasoning attacks. (2) Real-time deployments that combine self-consistency tests with lightweight instance selection where cost is the binding constraint. (3) Broader benchmarks---cross-domain, multilingual, and standardized attack suites (PoisonedRAG, FlippedRAG)---to stress-test defense layers. (4) Theoretical bounds on patch efficacy, connecting LOO filtering to the geometry of poison coalitions.

\subsection{Broader Impact}
\label{sec:impact}

Robust RAG matters most in high-stakes applications such as medicine, finance, and law \citep{ram2023incontext}. Increased robustness shifts, rather than ends, the attacker--defender dynamic; by open-sourcing the patches and stress-testing pipelines, the community can probe these defenses before attackers do. Our poison-generation code is released for reproducibility of the defense evaluation; the poisons are simple LLM rewrites with no capability beyond what current attack literature already documents \citep{zou2024poisonedrag, pan2023misinformation}.

\section{Conclusion}
\label{sec:conclusion}

We introduced RAGuard, a modular two-layer defense that strengthens RAG systems against factual corpus poisoning. The first layer fine-tunes dense retrievers on synthetic poisons (fabrications, contradictions, reasoning traps) to downrank malicious content; the second layer, \zkip{}, filters retrieved passages by measuring the causal influence of each on the generated answer---without poison labels, gold answers, or model internals. On poisoned NQ, \zkip{} eliminates measured attack success in every defended configuration while keeping retrieval quality within 0.03 of the clean-corpus baseline, at a cost of $k{+}1$ generator passes per query; supervised analyses on NQ and BEIR confirm that its counterfactual signals carry learnable poison structure. The clearest open problems are coordinated multi-poison attacks, standardized attack baselines, and reducing inference overhead; we release code, datasets, and evaluation harnesses to support that work.

\section*{Limitations}
\label{sec:limitations}

\paragraph{Computational overhead.}
\zkip{} requires $k{+}1$ generator passes per query ($6\times$ at $k{=}5$). Batching reduces wall-clock cost on local models, but for API-priced or latency-critical deployments the overhead is substantial. \zkip{} is most practical for high-stakes, lower-throughput pipelines or selective application to flagged queries.

\paragraph{Single-document removal misses coordinated attacks.}
If multiple poisoned documents reinforce one another, removal of any single item may not restore the correct response, muting the LOO signal (\S\ref{sec:multipoison}, \S\ref{sec:failures}). Iterative removal raises cost and only partially closes this gap; we have not yet measured the multi-poison frontier.

\paragraph{False positives.}
Filtering on output change can flag benign but opinionated or out-of-distribution documents, especially for ambiguous queries or factual disagreements; this is the source of the Recall@5 cost we observe. Integrating additional heuristics or weak supervision could reduce such false positives.

\paragraph{Evaluation scope.}
Our defended (\zkip{}) evaluation covers NQ at 10\% and 30\% poisoning with a single seed per condition; the undefended grid covers 5--30\%. Metrics derive from a limited set of runs constrained by compute, and the number of attackable queries per condition is modest (33--190), which explains the non-monotonic ASR pattern across ratios. ASR 0.000 means no successful attacks \emph{among the attacks and samples we tested}, not a guarantee. On BEIR we report poisoning impact and supervised detection results (Appendices~\ref{app:fullresults} and~\ref{app:classifier}); completing the \zkip{}-defended BEIR grid, adding seeds, and comparing against standardized attacks (PoisonedRAG, FlippedRAG) and recent defense baselines are the highest-priority evaluation extensions.

\paragraph{Effectiveness is strongest for factual QA.}
The stability signal presumes a question with an (approximately unique) factual answer. For subjective, multi-hop, or generative tasks, answer variation under context removal is expected even without poisoning, and \zkip{}'s signals weaken (Appendix~\ref{app:threat}).

\paragraph{Threat model is implicitly dense-retrieval-only.}
Our poisons are LLM rewrites that alter semantics while largely preserving the original keywords: BM25 shows ASR 0.000 at 5--20\% poisoning (0.011 at 30\%) against them with no defense, because poisoned and gold documents share the query's terms and term-frequency ranking can rarely distinguish them. The threat model is therefore optimized for dense embeddings and does not reflect realistic poisoning of hybrid lexical+semantic pipelines. A real adversary would craft poisons to evade both signals simultaneously; evaluating RAGuard against such hybrid-targeting poisons (our released \texttt{--hard\_mode} generator produces keyword-substituted variants) is required future work.

\paragraph{Adversarially trained retriever may be distribution-fragile.}
Fine-tuning on a fixed distribution of synthetic poisons risks two failure modes: (i) \emph{overfitting} to the three LLM-rewrite families used during training, such that gradient-optimized or domain-shifted attack passages are not downranked; and (ii) \emph{degraded generalization} to unseen query--document pairs from out-of-distribution domains. On BEIR (NFCorpus), a medical corpus distinct from NQ, the adversarially trained retriever's ASR is comparable to the clean model's (Table~\ref{tab:beir-full}), and BEIR's low absolute retrieval numbers make strong conclusions impossible. We do not claim the trained retriever generalizes to unseen domains, and we rely on \zkip{} as the attack-agnostic safety net. Three mitigations worth exploring are: (a) \emph{diverse poison generation}---including gradient-optimized passages \citep{su2024corpus} and backdoor-trigger rewrites \citep{long2025backdoor} in training, not only LLM rewrites; (b) \emph{periodic retraining} on newly discovered attack corpora as the threat landscape evolves; and (c) \emph{data-augmentation via domain mixing}, training on joint NQ+BEIR+open-domain triples to reduce query-distribution sensitivity.

\bibliographystyle{plainnat}
\bibliography{custom}

\appendix

\section{Full Baseline and \zkip{} Evaluation Results}
\label{app:fullresults}

We report full retrieval/defense results across clean, poisoned, and \zkip{}-defended settings in Tables~\ref{tab:nq-full} and~\ref{tab:beir-full}. Labels follow \S\ref{sec:setup}: ``Dense (adv-trained)'' is the retriever fine-tuned on synthetic poisons; ``Dense (clean)'' is trained on clean triples only. Every value is taken directly from the released evaluation artifacts in the code repository; conditions without a completed run are omitted rather than estimated.

\begin{table}[h]
\centering
\small
\begin{tabular}{llccc}
\toprule
\textbf{Setting} & \textbf{Retriever} & \textbf{Recall@5} & \textbf{MRR} & \textbf{ASR}~$\downarrow$ \\
\midrule
\multicolumn{5}{l}{\textit{Clean baseline}} \\
NQ & BM25                    & 0.068 & 0.053 & 0.000 \\
NQ & Dense (clean)           & 0.282 & 0.200 & 0.000 \\
\midrule
\multicolumn{5}{l}{\textit{Adversarially trained, evaluated on clean corpus}} \\
NQ & Dense (adv-trained)     & 0.301 & 0.201 & 0.000 \\
\midrule
\multicolumn{5}{l}{\textit{Under poisoning attack}} \\
NQ (5\%)  & BM25                  & 0.070 & 0.055 & 0.000 \\
NQ (5\%)  & Dense (clean)         & 0.268 & 0.181 & 0.061 \\
NQ (5\%)  & Dense (adv-trained)   & 0.323 & 0.220 & 0.091 \\
NQ (10\%) & BM25                  & 0.070 & 0.056 & 0.000 \\
NQ (10\%) & Dense (clean)         & 0.259 & 0.176 & 0.101 \\
NQ (10\%) & Dense (adv-trained)   & 0.319 & 0.215 & 0.072 \\
NQ (20\%) & BM25                  & 0.075 & 0.059 & 0.000 \\
NQ (20\%) & Dense (clean)         & 0.256 & 0.179 & 0.029 \\
NQ (20\%) & Dense (adv-trained)   & 0.327 & 0.218 & 0.065 \\
NQ (30\%) & BM25                  & 0.071 & 0.059 & 0.011 \\
NQ (30\%) & Dense (clean)         & 0.256 & 0.175 & 0.053 \\
NQ (30\%) & Dense (adv-trained)   & 0.318 & 0.212 & 0.068 \\
\midrule
\multicolumn{5}{l}{\textit{Poisoning + \zkip{} defense}} \\
NQ (10\%) & BM25 + \zkip{}                & 0.070 & 0.056  & \textbf{0.000} \\
NQ (10\%) & Dense (clean) + \zkip{}       & 0.259 & 0.176  & \textbf{0.000} \\
NQ (10\%) & Dense (adv-trained) + \zkip{} & 0.314 & 0.215  & \textbf{0.000} \\
NQ (30\%) & BM25 + \zkip{}                & 0.071 & 0.059  & \textbf{0.000} \\
NQ (30\%) & Dense (clean) + \zkip{}       & 0.256 & 0.175  & \textbf{0.000} \\
NQ (30\%) & Dense (adv-trained) + \zkip{} & 0.274 & 0.196  & \textbf{0.000} \\
\bottomrule
\end{tabular}
\caption{Natural Questions retrieval/defense results across clean, poisoned, and \zkip{}-defended settings. \zkip{} eliminates measured ASR in every defended configuration. Recall@5 and MRR are computed over the full 1{,}000-query set; ASR is computed over the queries where a poisoned document is present in the retrieved candidates (33, 69, 139, and 190 queries at 5\%, 10\%, 20\%, and 30\% poisoning respectively). Defended runs exist at 10\% and 30\%; the 20\% defended condition has not been run and is omitted rather than estimated.}
\label{tab:nq-full}
\end{table}

\begin{table}[h]
\centering
\small
\begin{tabular}{llccc}
\toprule
\textbf{Setting} & \textbf{Retriever} & \textbf{Recall@5} & \textbf{MRR} & \textbf{ASR}~$\downarrow$ \\
\midrule
\multicolumn{5}{l}{\textit{Clean baseline}} \\
BEIR & BM25            & 0.018 & 0.013 & 0.000 \\
BEIR & Dense (clean)   & 0.019 & 0.014 & 0.000 \\
\midrule
\multicolumn{5}{l}{\textit{Adversarially trained, evaluated on clean corpus}} \\
BEIR & Dense (adv-trained) & 0.018 & 0.013 & 0.000 \\
\midrule
\multicolumn{5}{l}{\textit{Under poisoning attack (30\% poisoned corpus)}} \\
BEIR (30\%) & BM25                & 0.016 & 0.013 & 0.000 \\
BEIR (30\%) & Dense (clean)       & 0.021 & 0.015 & 0.008 \\
BEIR (30\%) & Dense (adv-trained) & 0.019 & 0.014 & 0.007 \\
\bottomrule
\end{tabular}
\caption{BEIR (NFCorpus) retrieval results on the clean and 30\%-poisoned corpora (2{,}481 ASR-eligible queries). Absolute retrieval numbers are low for all systems on NFCorpus; the relevant observations are relative: dense ASR is an order of magnitude above BM25's (which stays at 0.000), and adversarial training does not meaningfully reduce ASR on this out-of-domain corpus. \zkip{}-defended BEIR runs are not yet complete and are deliberately not reported (\S\ref{sec:limitations}); supervised detection results on BEIR appear in Appendix~\ref{app:classifier}.}
\label{tab:beir-full}
\end{table}

\section{Threat Model Scope}
\label{app:threat}

\paragraph{Attack classes in scope.}
RAGuard is designed and evaluated against \emph{factual poisoning}: an adversary injects passages containing false factual claims---fabricated facts, contradictions of the gold evidence, and corrupted reasoning chains---into the retrieval corpus, with the goal of changing the answers a RAG pipeline produces. The adversary is assumed to have write access to the corpus (e.g., via user-generated content, wiki edits, or web crawl injection) but \emph{no} access to model weights, query inputs, or the generation prompt. Within this class, both retrieval-level attacks (the poisoned passage ranks above the gold passage) and generation-level attacks (the poisoned passage is retrieved alongside gold evidence and confuses the generator) are in scope: Layer 1 (adversarial retriever training) targets the former, and \zkip{} operates at the generation stage after retrieval, so it applies to both. All experiments in this paper instantiate this threat model with three poison families (fabrication, contradiction, reasoning trap) at corpus poison ratios of 5--30\%.

\paragraph{Attack classes out of scope.}
Several attack classes documented in the literature are explicitly \emph{not} covered by our evaluation. \emph{Prompt hijacking / instruction injection}---embedding instructions in documents to override system prompts---may be partially caught by \zkip{}, since injected instructions tend to destabilize the decoded answer, but we do not evaluate this and make no claim. \emph{Opinion manipulation} \citep{chen2025flippedrag} shifts model outputs on subjective questions with no single factual answer; \zkip{}'s answer-stability signal is less meaningful without a factual reference point, and we expect degraded performance. \emph{Retrieval jamming / denial-of-service}---flooding the corpus with irrelevant documents to suppress relevant results---is not addressed; \zkip{} can only filter what is retrieved. \emph{Backdoor-style attacks with trigger tokens} \citep{long2025backdoor} may be detected when the trigger causes answer instability, but a targeted backdoor engineered to produce a \emph{stable} wrong answer could evade detection (\S\ref{sec:failures}). \emph{Adversarial queries} are excluded: RAGuard assumes the query stream is clean. \emph{White-box gradient attacks} on the retriever or generator (e.g., Joint-GCG, \citealp{wang2025jointgcg}; greedy corpus poisoning, \citealp{su2024corpus}) are not the primary target; adversarial retriever training provides at most partial defense against them.

\paragraph{Corpus and query assumptions.}
RAGuard assumes (i) the query stream is clean (no adversarially crafted queries); (ii) the retrieval corpus is partially poisoned, with poison ratios of 5--30\% in our experiments; and (iii) the gold (correct) document exists somewhere in the corpus, even if not always retrieved. Assumption (iii) matters for \zkip{}: if no correct evidence is retrievable, removing a poison cannot restore a correct answer, and the stability signal degrades toward the failure mode described in \S\ref{sec:failures}.

\section{Retriever Training Details}
\label{app:training}

\begin{table}[h]
\centering
\small
\begin{tabular}{ll}
\toprule
\textbf{Component} & \textbf{Value} \\
\midrule
Base encoder & \texttt{all-MiniLM-L6-v2} \\
Pooling & Mean over last hidden state \\
Loss & Triplet margin loss, margin $=0.2$ \\
Optimizer & AdamW \\
Learning rate & $2\times10^{-5}$ \\
Batch size & 16 \\
Epochs & 3 \\
Max sequence length & 256 tokens \\
Poison ratio (training) & 0.30, even across 3 families \\
Negatives (poisoned triples) & The poisoned rewrite of the gold doc \\
Negatives (clean triples) & Random clean passage from corpus \\
Training files & \texttt{train\_triples\_nq\_poisoned.jsonl} \\
               & \texttt{train\_triples\_clean.jsonl} \\
\bottomrule
\end{tabular}
\caption{Full retriever fine-tuning configuration. ``Dense (clean)'' uses the clean triples file; ``Dense (adv-trained)'' uses the poisoned triples file with the same hyperparameters.}
\label{tab:trainconfig}
\end{table}

Each training triple is $\{$query, gold document, negative document$\}$. For triples carrying a poison label, the negative is the attack-family-specific rewrite of that triple's gold document, making the contrastive signal directly anti-poison; clean triples use in-corpus random negatives. The released training script (\texttt{retrievers/train.py}) additionally supports contrastive and in-batch-negative losses, hard-negative selection, a configurable training poison ratio, and per-epoch loss logging for reproducibility; the configuration in Table~\ref{tab:trainconfig} reproduces the checkpoints used in this paper. \zkip{} defaults: $\lambda = 0.5$, removal budget $m = 1$, $k = 5$, answer encoder \texttt{all-MiniLM-L6-v2}, generator GPT-4o-mini (temperature 0.2, top-$p$ 0.95, max 128 tokens).

\section{Supervised Poison Classification Ablations}
\label{app:classifier}

This appendix reports supervised poison-classification ablations that complement \zkip{}. Whereas \zkip{} is label-free and operates by counterfactual sensitivity, these ablations test whether (i) poisoned passages are distinguishable from gold passages directly from text, and (ii) \zkip{}-derived influence signals are sufficiently informative for a supervised model to predict poison labels. These results are a supervised upper bound: they require labeled poison data during training, while \zkip{} does not.

\subsection{Experimental Setup}
We extract eight features per retrieved document: \textbf{rank\_pos} (normalized retrieval rank), \textbf{sim\_q} (query--document cosine), \textbf{sim\_ans} (similarity between baseline and counterfactual answers), \textbf{entropy\_delta} ($\Delta H_i$ per Equation~\ref{eq:deltah}), \textbf{f1\_delta} and \textbf{em\_delta} (answer-score change on removal; gold-dependent, used only in this supervised analysis), \textbf{flip\_wrong\_to\_right} (indicator that removal corrects a wrong answer; gold-dependent, supervised analysis only), and \textbf{sim\_z} (embedding outlier z-score within the retrieved set). The released feature extractor exposes an \texttt{oracle\_free} mode that drops the gold-dependent features for inference-time use.

\subsection{Text-Level Classification on Gold vs.\ Poison Pairs}

\begin{table}[h]
\centering
\small
\begin{tabular}{llcccccc}
\toprule
\textbf{Dataset} & \textbf{Poison} & \textbf{Acc.} & \textbf{Prec.} & \textbf{Recall} & \textbf{F1} & \textbf{ROC-AUC} & \textbf{AUPRC} \\
\midrule
\multirow{4}{*}{NQ}
 & 5\%  & 0.594 & 0.851 & 0.228 & 0.360 & 0.685 & 0.719 \\
 & 10\% & 0.593 & 0.844 & 0.228 & 0.359 & 0.690 & 0.720 \\
 & 20\% & 0.592 & 0.843 & 0.226 & 0.356 & 0.686 & 0.720 \\
 & 30\% & 0.591 & 0.853 & 0.220 & 0.350 & 0.685 & 0.717 \\
\midrule
\multirow{2}{*}{BEIR}
 & 5\%  & 0.613 & 0.649 & 0.490 & 0.559 & 0.687 & 0.678 \\
 & 10\% & 0.612 & 0.657 & 0.466 & 0.546 & 0.689 & 0.674 \\
\bottomrule
\end{tabular}
\caption{BERT poison-classifier performance on gold vs.\ poison document pairs: high precision but low recall on NQ indicates many poisons are lexically subtle rewrites that text alone cannot flag without over-triggering.}
\label{tab:bertclf}
\end{table}

Across NQ, AUPRC remains stable at $\approx$0.717--0.720 over poisoning ratios from 5\% to 30\%, with ROC-AUC $\approx$0.685--0.690 (Table~\ref{tab:bertclf}). Precision is high ($\approx$0.84--0.85) but recall is low ($\approx$0.22--0.23): the classifier is conservative, flagging only a limited subset of poisons. Many poisons are semantically subtle rewrites that remain lexically and stylistically similar to gold documents. On BEIR, the profile is more balanced (precision $\approx$0.65, recall $\approx$0.47--0.49), suggesting poisoned passages are comparatively easier to separate at the document-text level there, potentially due to greater topical diversity or larger stylistic shifts.

\subsection{Feature-Level Classification Using \zkip{} Influence Signals}

\begin{table}[h]
\centering
\small
\begin{tabular}{lcc}
\toprule
\textbf{Metric} & \textbf{LogReg} & \textbf{BERT} \\
\midrule
AUPRC       & 0.377 & 0.732 \\
ROC-AUC     & 0.571 & 0.814 \\
F1          & 0.514 & 0.636 \\
Precision   & 0.442 & 0.531 \\
Recall      & 0.614 & 0.792 \\
Specificity & 0.536 & 0.581 \\
\bottomrule
\end{tabular}
\caption{Classifier performance on labeled NQ using \zkip{}-derived influence features: nonlinear models nearly double AUPRC over a linear baseline, indicating exploitable interactions among counterfactual signals.}
\label{tab:featclf}
\end{table}

A logistic-regression baseline achieves AUPRC 0.377 and ROC-AUC 0.571, while the BERT-based classifier improves substantially to AUPRC 0.732 and ROC-AUC 0.814, with F1 increasing from 0.514 to 0.636 (Table~\ref{tab:featclf}). The AUPRC gain matters under class imbalance and indicates that nonlinear models exploit interactions among influence features (e.g., combining answer destabilization with retrieval rank and similarity cues) that a linear model cannot capture. In the logistic-regression feature-importance analysis, retrieval rank (\textbf{rank\_pos}) dominates, followed by query similarity (\textbf{sim\_q}) and answer stability (\textbf{sim\_ans}), with the entropy differential contributing a smaller but nonzero share.

\subsection{Takeaway and Relationship to \zkip{}}
These supervised results support the central motivation of \zkip{}: poisoned passages induce measurable, systematic changes in generation under counterfactual removal, and these changes contain learnable signal. However, unlike \zkip{}, supervised classifiers require poison labels for training and may not generalize to unseen attack styles without continual relabeling and retraining. We therefore treat the learned classifiers as ablations that validate the informativeness of \zkip{}'s influence signals, while maintaining \zkip{} as the primary, label-free defense.

\end{document}